\documentclass[letterpaper, 10 pt, conference]{ieeeconf/ieeeconf}
\IEEEoverridecommandlockouts                            
\usepackage{cite}
\usepackage{amsmath}
\usepackage{amssymb}
\usepackage{breqn}
\usepackage{float}
\usepackage{tabularx}
\usepackage{graphicx,dblfloatfix}
\usepackage{siunitx}
\usepackage{textcomp}
\usepackage{bbm}
\usepackage{dsfont}
\usepackage[dvipsnames]{xcolor}
\usepackage{soul}
\soulregister\cite7
\soulregister\ref7
\soulregister\pageref7
\usepackage[symbol]{footmisc}

\DeclareMathOperator*{\argmax}{arg\,max} 
\DeclareMathOperator*{\truemax}{max} 
\newcommand{\x}{\mathbf{x}}

\usepackage{color, colortbl}
\definecolor{Gray}{gray}{0.85}
\definecolor{LightCyan}{rgb}{0.88,1,1}
\newcolumntype{a}{>{\columncolor{Gray}}c}
\newcolumntype{b}{>{\columncolor{LightCyan}}c}

\title{\LARGE \bf
Information-Guided Robotic Maximum Seek-and-Sample \\ in Partially Observable Continuous Environments
}

\author{Genevieve~Flaspohler,*$^{1,2}$ Victoria~Preston,*$^{1,2}$ Anna~P.M.~Michel,$^{2}$ Yogesh~Girdhar,$^{2}$ and~Nicholas~Roy$^{1}$%
\thanks{*These authors contributed equally to the paper.}
\thanks{$^{1}$CSAIL, Massachusetts Institute of Technology, Cambridge MA, USA
        {\tt \small [geflaspo,vpreston,nickroy]@csail.mit.edu}}%
\thanks{$^{2}$Applied Ocean Physics and Engineering, Woods Hole Oceanographic Institution, Woods Hole MA, USA
        {\tt\small [amichel,yogi]@whoi.edu}}%
}

\begin{document}

\maketitle

\begin{abstract}
We present PLUMES, a planner for localizing and collecting samples at the global maximum of an \textit{a priori} unknown and partially observable continuous environment. This ``maximum seek-and-sample'' (MSS) problem is pervasive in the environmental and earth sciences. Experts want to collect scientifically valuable samples at an environmental maximum (e.g., an oil-spill source), but do not have prior knowledge about the phenomenon's distribution. We formulate the MSS problem as a partially-observable Markov decision process (POMDP) with continuous state and observation spaces, and a sparse reward signal. To solve the MSS POMDP, PLUMES uses an information-theoretic reward heuristic with continuous-observation Monte Carlo Tree Search to efficiently localize and sample from the global maximum. In simulation and field experiments, PLUMES collects more scientifically valuable samples than state-of-the-art planners in a diverse set of environments, with various platforms, sensors, and challenging real-world conditions.
\end{abstract}


\section{Introduction}
\label{sec:intro}
\begin{figure}[t]
\centering
\includegraphics[width=0.95\linewidth]{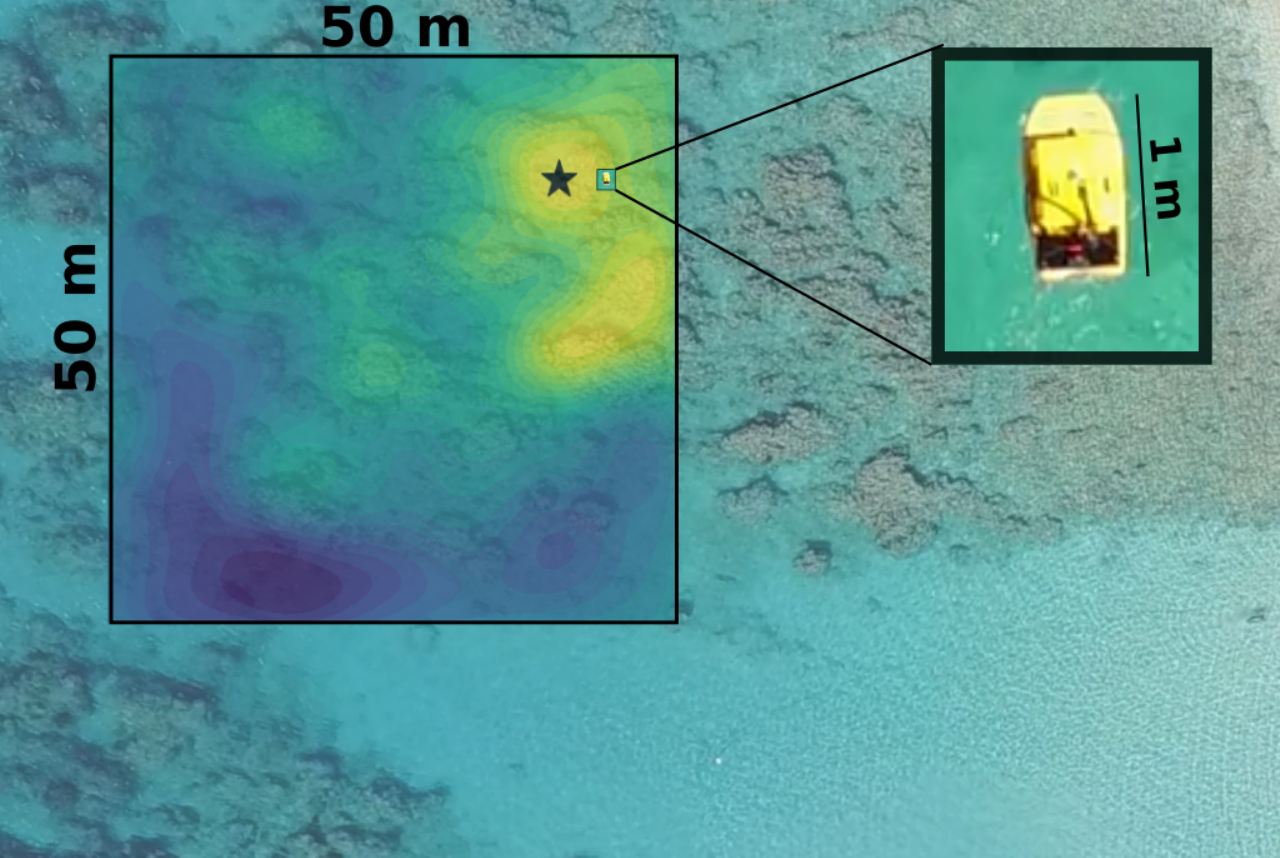}
\caption{\textbf{Coral head localization with an autonomous surface vehicle (ASV):} The objective of the ASV is to find and sample at the most exposed (shallowest) coral head in a region of Bellairs Fringing Reef, Barbados. Overlaid on the aerial photo is the \textit{a priori} unknown bathymetry of the region (yellow is shallow, blue is deep). Equipped with an acoustic point altimeter, the ASV must explore to infer the location of the maximum (marked with a star) and then sample at that coral colony. \vspace{-0.6cm}}
\label{fig:figure1_coral}
\end{figure}

In many environmental and earth science applications, experts want to collect scientifically valuable samples of a maximum (e.g., an oil spill source), but the distribution of the phenomenon is initially unknown. This \textit{maximum seek-and-sample} (MSS) problem is pervasive. Canonically, samples are collected at predetermined locations by a technician or by a mobile platform following a uniform coverage trajectory. These non-adaptive strategies result in sample sparsity at the maximum and may be infeasible when the geometric structure of the environment is unknown (e.g., boulder fields) or changing (e.g., tidal zones). Increasing the number of valuable samples at the maximum requires adaptive online planning and execution. We present PLUMES --- \textbf{P}lume \textbf{L}ocalization under \textbf{U}ncertainty using \textbf{M}aximum-Valu\textbf{E} information and \textbf{S}earch --- an adaptive algorithm that enables a mobile robot to efficiently localize and densely sample an environmental maximum, subject to practical challenges including dynamic constraints, unknown geometric map and obstacles, and noisy sensors with limited field-of-view. Fig.~\ref{fig:figure1_coral} shows a motivating application: coral head localization.

\textbf{Informative Path Planning:} The MSS problem is closely related to informative path planning (IPP) problems. Canonical offline IPP techniques for pure information-gathering that optimize submodular coverage objectives can achieve near-optimal performance \cite{Srinivas2012, binney2012branch}. However, in the MSS problem, the value of a sample depends on the unknown maximum location, requiring adaptive planning to enable the robot to select actions that explore to localize the maximum and then seamlessly transition to selecting actions that exploitatively collect valuable samples there. Even for adaptive IPP methods, the MSS problem presents considerable challenges. The target environmental phenomenon is partially observable and most directly modeled as a continuous scalar function. Additionally, efficient maximum sampling with a mobile robot requires consideration of vehicle dynamics, travel cost, and a potentially unknown obstacle map. Handling these challenges in combination excludes adaptive IPP algorithms that use discrete state spaces \cite{Lim2016, Arora2017}, known metric maps \cite{singh2009nonmyopic, Jawaid2015}, or unconstrained sensor placement \cite{Krause2008}.

\textbf{The MSS POMDP:} Partially-observable Markov decision processes (POMDPs) are general models for decision-making under uncertainty that allow the challenging aspects of the MSS problem to be encoded. We define the MSS POMDP, in which the partially observable state represents the continuous environmental phenomenon and a sparse reward function encodes the MSS scientific objective by giving reward only to samples sufficiently close to the global maximum. Solving a POMDP exactly is generally intractable, and the MSS POMDP is additionally complicated by both continuous state and observation spaces, and the sparse MSS reward function. This presents the two core challenges that PLUMES addresses: performing online search in a belief-space over continuous functions, and overcoming reward function sparsity.

\vspace{0.1cm}
\textbf{Planning over Continuous Domains:} In the MSS problem, the state of the environment can be modeled as a continuous function. PLUMES uses a Gaussian Process (GP) model to represent the belief over this continuous function, and must plan over the uncountable set of possible GP beliefs that arise from future continuous observations. To address planning in continuous spaces, state-of-the-art online POMDP solvers use deterministic discretization \cite{ling2016gaussian} or a combination of sampling techniques and particle filter belief representations \cite{somani2013despot, kurniawati2016online, Silver2010, sunberg2017value}. Efficiently discretizing or maintaining a sufficiently rich particle set to represent the underlying continuous function in MSS applications is itself a challenging problem, and can lead to inaccurate inference of the maximum \cite{dallaire2009bayesian}. Other approaches have considered using the maximum-likelihood observation to make search tractable \cite{Marchant2014a}. However, this assumption can compromise search and has optimality guarantees only in linear-Gaussian systems \cite{platt2010belief}. Instead, PLUMES uses Monte Carlo Tree Search (MCTS) with progressive widening, which we call \textit{continuous-observation MCTS}, to limit planning tree growth \cite{couetoux2011continuous} and retain optimality \cite{auger2013continuous} in continuous environments.

\vspace{0.1cm}
\textbf{Rewards and Heuristics:} In the MSS POMDP, the reward function is sparse and does not explicitly encode the value of exploration. Planning with sparse rewards requires long-horizon information gathering and is an open problem in robotics \cite{smart2002effective}. To alleviate this difficulty, less sparse heuristic reward functions can be optimized in place of the true reward, but these heuristics need to be selected carefully to ensure the planner performs well with respect to the true objective. In IPP, heuristics based on the value of information have been applied successfully \cite{Marchant2014a,Sun2017,Krause2008,Hitz2017}, primarily using the GP-UCB criteria \cite{Contal2013,Srinivas2012}. We demonstrate that within practical mission constraints, using UCB as the heuristic reward function for the MSS POMDP can lead to suboptimal convergence to local maxima due to a mismatch between the UCB heuristic and the true MSS reward. Instead, PLUMES takes advantage of a heuristic function from the Bayesian optimization (BO) community for state-of-the-art black-box optimization \cite{wang2017max}, which we call \textit{maximum-value information} (MVI). MVI overcomes sparsity and encourages long-term information gathering, while still converging to the true reward of the MSS POMDP.

\vspace{0.1cm}
The contribution of this paper is the MSS POMDP formalism and the corresponding PLUMES planner, which by virtue of its belief model, information-theoretic reward heuristic, and search framework, enables efficient maximum seek and sample with asymptotic optimality guarantees in continuous environments. PLUMES extends the state-of-the-art in MSS planners by applying a BO heuristic reward function to MSS that alleviates the challenges of the true sparse MSS reward function, and integrating GP belief representations within continuous-observation MCTS. The utility of PLUMES for MSS applications is demonstrated in extensive simulation and field trials,  showing a statistically significant performance improvement over state-of-the-art baselines.

\section{Maximum Seek-and-Sample POMDP}
\label{prob}
We formalize the MSS problem by considering a target environmental domain as a $d$-dimensional compact set $\mathbb{X}_w \subset \mathbb{R}^d$. We allow $\mathbb{X}_w$ to contain obstacles with arbitrary geometry and let $\mathbb{X} \subset \mathbb{X}_w$ be the set of reachable points with respect to the robot's initial pose. We assume there is an unknown underlying continuous function $f: \mathbb{X}_w \to \mathbb{R}$ representing the value of a continuous phenomenon of interest. The objective is to find the unique global maximizer $\x^* = \argmax_{\x \in \mathbb{X}} f(\x)$ by safely navigating while receiving noisy observations of this function $f$. Because $f$ is unknown, we cannot access derivative information or any analytic form. 

We model the process of navigating and generating observations as the MSS POMDP: an 8-tuple $(\mathcal{S}, \mathcal{A}, \mathcal{Z}, T, O, R, \gamma, b_0)$:
\begin{itemize}
    \item $\mathcal{S}$: continuous state space of the robot and environment
    \item $\mathcal{A}$: discrete set of action primitives
    \item $\mathcal{Z}$: continuous space of possible observations
    \item $T$: $\mathcal{S} \times \mathcal{A} \to \mathcal{P}(\mathcal{S})$, the transition function, i.e., \\ $\mathbf{Pr}(S_{t+1} = s' \mid S_{t} = s, A_t = a)$
    \item $O$: $\mathcal{S} \times \mathcal{A} \to \mathcal{P}(\mathcal{Z})$, the observation model, i.e., \\ $\mathbf{Pr}(Z_{t+1} = z \mid S_{t+1} = s, A_t = a)$
    \item $R$: $\mathcal{S} \times \mathcal{A} \to \mathbb{R}$, the reward of taking action $a$ when robot's state is $s$, i.e., $R(s, a)$
    \item $\gamma$: discount factor, $0 \leq \gamma \leq 1$
    \item $b_0$: initial belief state of the robot, $b_0 \in \mathcal{P}(S_0)$
\end{itemize}
where $\mathcal{P}(\cdot)$ denotes the space of probability distributions over the argument.

The Bellman equation is used to recursively quantify the value of belief $b_t = \mathcal{P}(S_t)$ over a finite horizon $h$ under policy $\pi: b_t \to a_t$ as: 
\begin{equation}
\begin{split}
  V^{\pi}_h(b_t) = &  \mathbb{E}[R(s_t, \pi(b_t))] + \\ 
  & \gamma \int_{z \in \mathcal{Z}} V^{\pi}_{h-1}(b^{\pi(b_t), z}_{t+1}) \text{ }\textbf{Pr}(z \mid b_t, \pi(b_t)) \text{ } \mathrm{d}z,
  \label{eq:value}
\end{split}
\end{equation}
where the expectation is taken over the current belief and $b^{\pi(b_t), z}_{t+1}$ is the updated belief after taking action $\pi(b_t)$ and observing $z \in \mathcal{Z}$. The optimal policy $\pi_h^*$ over horizon-$h$ is the maximizer of the value function over the space of possible policies $\Pi$: $\pi_h^* = \argmax_{\pi \in \Pi} V^{\pi}_h(b_t)$. However, Eq. \ref{eq:value} is intractable to compute in continuous state and observation spaces; the optimal policy must be approximated. PLUMES uses a receding-horizon, online POMDP planner and heuristic reward function to approximately solve the MSS POMDP in real-time on robotic systems. 

\vspace{0.5cm}
\section{THE PLUMES ALGORITHM}
\label{sec:model}
PLUMES is an online planning algorithm with a sequential decision-making structure:
 \begin{enumerate}
   \item Conditioned on $b_t$, approximate the optimal policy $\pi_h^*$ for finite horizon $h$ and execute the action $a = \hat{\pi}_h^*(b_t)$.
   \item Collect observations $z \in \mathcal{Z}$, according to $O$.
   \item Update $b_t$ to incorporate this new observation; repeat.
 \end{enumerate}
In the following sections, we define the specific choice of belief model, planning algorithm, and heuristic reward function that PLUMES uses to solve the MSS POMDP. 

\subsection{Gaussian Process Belief Model}
\label{sec:gp}
We assume the robot's pose $\x_t$ at planning iteration $t$ is fully observable, and the unknown environmental phenomenon $f$ is partially observable. The full belief-state is represented as a tuple $b_t$ of robot state $\x_t$ and environment belief $g_t = \mathcal{P}(f)$ at time $t$. Because $f$ is a continuous function, we cannot represent the belief $g_t$ as a distribution over discrete states, as is standard in POMDP literature \cite{kaelbling1998planning}, and must choose an alternate representation. PLUMES uses a Gaussian process (GP) \cite{Rasmussen2004} to represent $g_t$ conditioned on a history of past observations. This GP is parameterized by mean $\mu(\mathbf{x})$ and covariance function $\kappa(\mathbf{x}, \mathbf{x}')$.

As the robot traverses a location $\mathbf{x}$, it gathers observations $z \in \mathcal{Z}$ of $f$ subject to sensor noise $\sigma_n^2$, such that $z = f(\mathbf{x}) + \epsilon$ with $\epsilon \stackrel{i.i.d.}{\sim} \mathcal{N}(0, \sigma_n^2)$. Given a history $\mathcal{D}_t = \{\mathbf{x}_i, z_i\}_{i = 0}^D$ of $D$ observations and observation locations at planning iteration $t$, the posterior belief at a new location $\mathbf{x}' \in \mathbb{X}$ is computed: 
\begin{align}
    &g_t (\x') \mid {\mathcal{D}_t} \sim \mathcal{N}(\mu_{t}(\x'), \sigma_{t}^2(\x')), \text {where} \\
    &\mu_{t}(\x') = \kappa_t(\x')^\top(\mathbf{K}_t + \sigma_n^2\mathbf{I})^{-1} \mathbf{z}_t,    \label{eq:gp_mean} \\
    &\sigma_t^2(\x') = \kappa(\x', \x') - \kappa_t(\x')^\top(\mathbf{K}_t + \sigma_n^2\mathbf{I})^{-1}\kappa_t(\x'), \label{eq:gp_var}
\end{align}
where $\mathbf{z}_t = [z_0, \dots, z_{D-1}]^\top$, $\mathbf{K}_t$ is the positive definite kernel matrix with $\mathbf{K}_t[i, j] = \kappa(\x_i, \x_j)$ for all $\x_i, \x_j \in \mathcal{D}_t$, and $\kappa_t(\x') = [\kappa(\x_0, \x'), \dots, \kappa(\x_{D-1}, \x')]^\top$. 

\subsection{Planning with Continuous-Observation MCTS}
\label{sec:mcts}
PLUMES selects high-reward actions with receding-horizon search over possible belief states. This search requires a simulator that can sample observations and generate beliefs given a proposed action sequence. For PLUMES, this simulator is the GP model, which represents the belief over the continuous function $f$, and in turn simulates continuous observations from proposed action sequences by sampling from the Gaussian distribution defined by Eq.~\ref{eq:gp_mean} \& \ref{eq:gp_var}.

\begin{figure}[t!]
\centering
\vspace{0.2cm}
\includegraphics[width=0.85\linewidth]{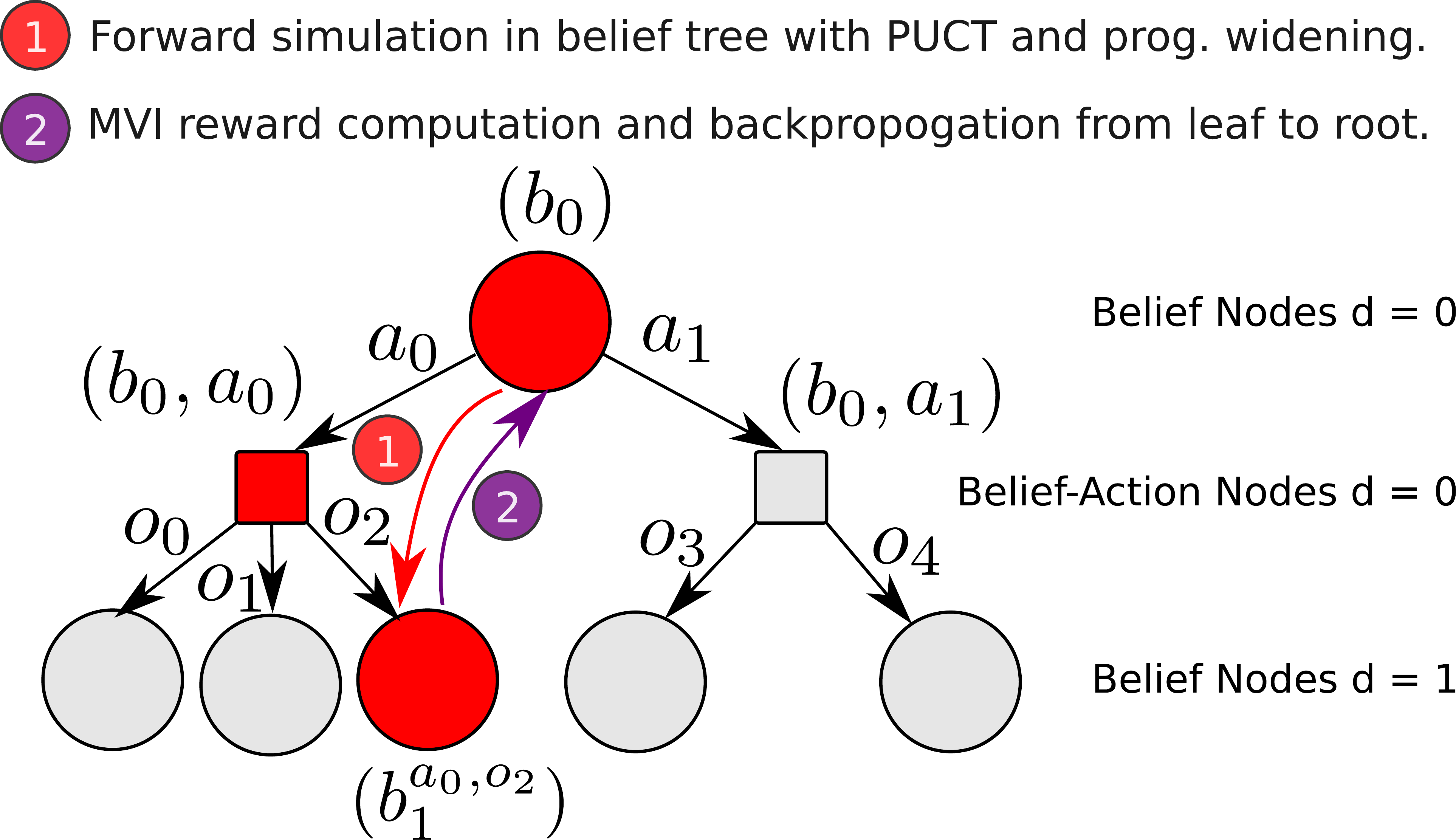}\hfill
\caption{\textbf{Continuous-observation MCTS:} Illustrated to horizon $h=1$, the tree consists of alternating belief and belief-action nodes. Action decisions are made at belief nodes and random belief transitions according to the observation function occur at belief-action nodes. Note that belief-action nodes have a varying number of children due to progressive widening and unequal simulation (not visualized) due to PUCT policy.}
\label{fig:figure2_mcts}
\vspace{-0.5cm}
\end{figure}

PLUMES uses continuous-observation MCTS to overcome the challenges of planning in continuous state and observation spaces. Continuous-observation MCTS has three stages: \textit{selection}, \textit{forward simulation}, and \textit{back-propagation}. Each node in the tree can be represented as the tuple of robot pose and GP belief, $b_t$ = \{$\x_t$, $g_t$\}. Additionally, we will refer to two types of nodes: belief nodes and belief-action nodes. The root of the tree is always a belief node, which represents the entire history of actions and observations up through the current planning iteration. Through selection and simulation, belief and belief-action nodes are alternately added to the tree (Fig.~\ref{fig:figure2_mcts}).

From the root, a rollout begins with the \textit{selection} stage, in which a belief-action child is selected according to the Polynomial Upper Confidence Tree (PUCT) policy \cite{auger2013continuous}. The PUCT value $\hat{Q}^*_{aug}(b_t, a)$ is the sum of the average heuristic rewards (i.e., MVI) from all previous simulations and a term that favors less-simulated action sequences:
\begin{equation}
  \hat{Q}^*_{aug}(b_t, a) = \hat{Q}^*(b_t, a) + \sqrt{\frac{{N(b_t)}^{e_d}}{N(b_t, a)}},
\end{equation}  
where $\hat{Q}^*(b_t, a)$ is the average heuristic reward of choosing action $a$ with belief $b_t$ in all previous rollouts, $N(b_t)$ is the number of times the node $b_t$ has been simulated, $N(b_t, a)$ is the number of times that particular action from node $b_t$ has been selected, and $e_d$ is a depth-dependent parameter\footnote{Refer to Table 1 of Auger et al. \cite{auger2013continuous} for parameter settings. \label{ft:auger}}.

Once a child belief-action node is selected, the action associated with the child is \textit{forward simulated} using the generative observation model $O$, and a new belief node is generated $b_{t+1} = \{\x_{t+1}, g_{t+1}\}$ as though the action were taken and samples observed. The simulated observations are drawn from the belief-action node's GP model $g_{t}$, and the robot's pose is updated deterministically based on the selected action. Since the observations in a GP are continuous, every sampled observation is unique with probability one. \textit{Progressive widening}, with depth-dependent parameter$^*$ $\alpha_d$ incrementally grows the tree by limiting the number of belief children of each belief-action node. When growing the tree, $b_{t+1}$ is either chosen to be the least visited node if $\lfloor N(b_t, a)^{\alpha_d} \rfloor=\lfloor (N(b_t, a) - 1)^{\alpha_d} \rfloor$, or otherwise is a new child with observations simulated from $b_t$. By limiting the width of the search tree and incrementally growing the number of explored children, progressive widening avoids search degeneracy in continuous environments.

Once a sequence of actions has been rolled out to a horizon $h$, the accumulated heuristic reward is \textit{propagated upward} from the leaves to the tree root. The average accumulated heuristic reward and number of queries are updated for each node visited in the rollout. Rollouts continue until the computation budget is exhausted. The most visited belief-action child of the root node is executed.

\begin{figure*}[b!]
\centering
\includegraphics[width=\linewidth]{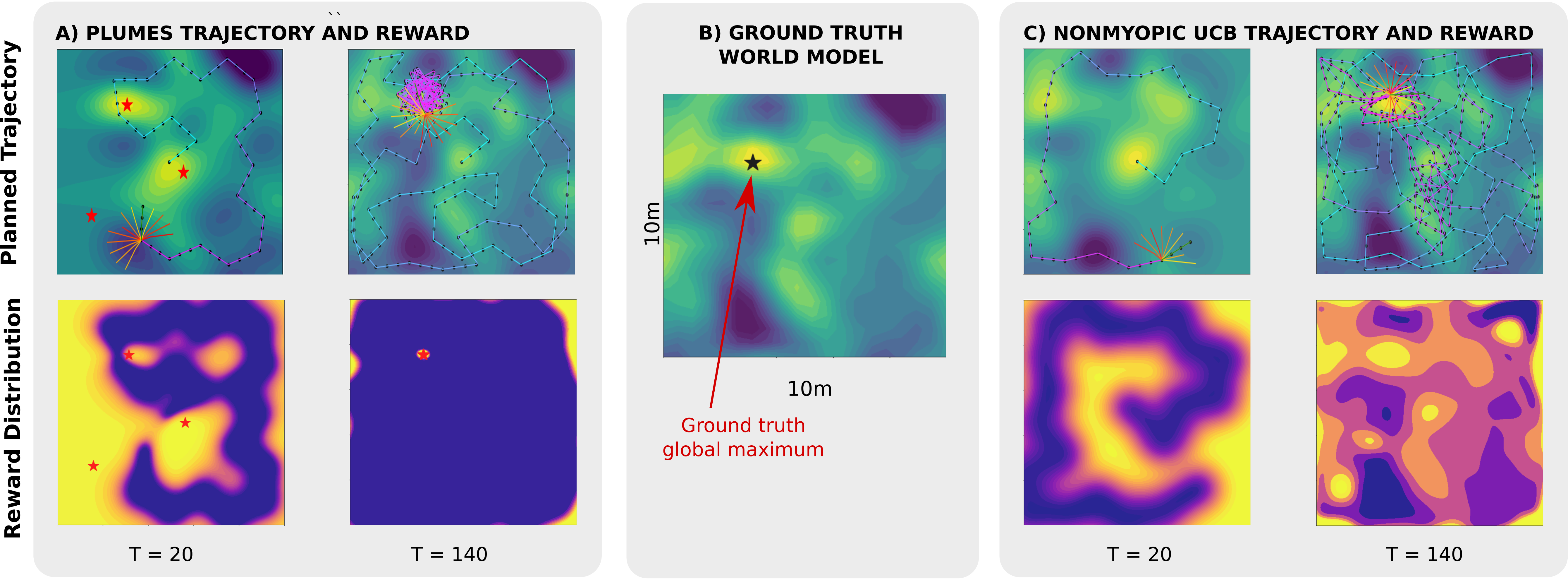}
\caption{\textbf{Convergence of MVI vs UCB heuristic:} The true environmental phenomenon with the global maximum marked by a star is shown in the center; high regions are colored yellow and low regions blue. In (A,C), the robot trajectory and corresponding reward functions are shown early (20 actions) and later (140 actions) in a mission. On the top row, snapshots of the robot belief state with planned trajectories are shown, with recent actions colored pink and earlier actions colored blue. Red stars mark maxima sampled by MVI. In the bottom row, the corresponding reward function is shown, with high-reward regions colored yellow and low reward regions colored purple.  By the end of the mission, MVI clearly converges to placing reward only at the global maximum, which in turn leads to efficient convergence of the robot. By contrast, the reward landscape resulting from canonically used UCB converges to the underlying function, causing the UCB planner to uniformly tour high-valued regions of the environment.}
\label{fig:figure3_mvi_conv}
\end{figure*}

Continuous-observation MCTS within PLUMES provides both practical and theoretical benefits. Practically, progressive-widening directly addresses search degeneracy by visiting belief nodes multiple times even in continuous observation spaces, allowing for a more representative estimate of their value. Theoretically, PLUMES can be shown to select asymptotically optimal actions. We briefly describe how analysis in Auger et al. \cite{auger2013continuous} for PUCT-MCTS with progressive widening in MDPs can be extended to PLUMES.  

Using standard methods \cite{kaelbling1998planning}, we can reduce the MSS POMDP to an equivalent belief-state MDP. This belief-state MDP has a state space equal to the set of all possible beliefs, and a transition distribution that captures the effect of both the dynamics and the observation model after each action. Planning in this representation is often intractable as the state space is continuous and infinite-dimensional. However, PLUMES plans directly in the belief-state MDP by using its GP belief state to compute the transition function efficiently.

Subsequently, Theorem 1 in Auger et al. \cite{auger2013continuous} shows that for an MDP with a continuous state space, like the belief-state MDP representation suggested, the value function estimated by continuous-observation MCTS asymptotically converges to that of the optimal policy:
\begin{equation}
\left\vert \hat{Q}_h^*(b_t, a)  - Q_h^*(b_t, a)\right\vert \leq \frac{C}{N(b_t, a)^{\gamma_d}},
\end{equation} 
with high probability \cite{auger2013continuous}, for constants $C > 0$ and $\gamma_d*$.

\subsection{Maximum-Value Information Reward}
\label{sec:mes}
The true state-dependent reward function for the MSS POMDP would place value on collecting sample points $\mathbf{x}$ within an $\epsilon$-ball of the true global maximum $\mathbf{x}^*$:
\begin{equation}
R(f, \mathbf{x}) = \mathds{1}_{\left \Vert \mathbf{x} - \mathbf{x}^* \right \Vert < \epsilon}, 
\label{eq:true_reward}
\end{equation}
where $\epsilon$ is determined by the scientific application. Optimizing this sparse reward function directly is challenging, so PLUMES approximates the true MSS reward by using the maximum-value information (MVI) heuristic reward \cite{wang2017max}. MVI initially encourages exploration behavior, but ultimately rewards exploitative sampling near the inferred maximum.

The belief-dependent MVI heuristic reward $\tilde{R}(b_t, \x)$ quantifies the expected value of having belief $b_t$ and collecting a sample at location $\x \in \mathbb{X}$. MVI reward quantifies the mutual information between the random variable $Z$, representing the observation at location $\x$, and $Z^*$, the random variable representing the value of the function $f$ at the global maximum:
\vspace{-0.25cm}
\begin{equation}
\tilde{R}(b_t, \x) = I(\{\x, Z\} ; Z^* \mid b_t), 
\label{eq:reward}
\end{equation}
where $Z^* = \truemax_{\x' \in \mathbb{X}} f(\x')$. To compute the reward of collecting a random observation $Z$ at location $\x$ under belief $b_t$, we approximate the expectation over the unknown $Z^*$ by sampling from the posterior distribution $z^*_i \sim p(Z^* \mid b_t)$ and use Monte Carlo integration with $M$ samples \cite{wang2017max}: 
\vspace{-0.25cm}
\begin{align}
\begin{split}
& \tilde{R}(b_t, \x)  =  H[\textbf{Pr}(Z \mid \x, b_t)] - \\
& \hspace{1.3cm} \mathbb{E}_{z' \sim \textbf{Pr}(Z^* \mid b_t)} [H[\textbf{Pr}(Z \mid  \x, b_t, Z^* = z')],
\end{split}\\
& \approx H[\textbf{Pr}(Z \mid \x, b_t)] - \frac{1}{M} \sum_{i = 0}^M H[\textbf{Pr}(Z \mid  \x, b_t, Z^*=z^*_i)].
\label{eq:final_mes}
\end{align}
Each entropy expression $H[\cdot]$ can be respectively approximated as the entropy of a Gaussian random variable with mean and variance given by the GP equations (Eq.~\ref{eq:gp_mean} \& \ref{eq:gp_var}), and the entropy of a truncated Gaussian, with upper limit $z^*_i$ and the same mean and variance.

To draw samples $z^*_i$ from the posterior $ p(Z^* \mid b_t)$, we employ spectral sampling \cite{rahimi2008random}. Spectral sampling draws a function $\hat{f}$, which has analytic form and is differentiable, from the posterior belief of a GP with stationary covariance function \cite{wang2017max,hernandez2014predictive}. To complete the evaluation of Eq.~\ref{eq:final_mes}, $z_i^* \sim p(Z^* \mid b_t)$ can be computed by applying standard efficient global optimization techniques (e.g., sequential least squares programming, quasi-Newton methods) to find the global maximum of the sampled $\hat{f}$. This results in the following expression for MVI reward \cite{wang2017max}:
\begin{equation}
\tilde{R}(b_t, \x) \approx \frac{1}{M} \sum_{i = 0}^M  \frac{\gamma_{z_i^*}(\x) \phi(\gamma_{z_i^*}(\x))}{2 \Phi(\gamma_{z_i^*}(\x))}-\text{log}(\Phi(\gamma_{z_i^*}(\x)))
\label{eq:wj_final}
\end{equation}
where $\gamma_{z_i^*}(\x) = \frac{z_i^* - \mu_t(\x)}{\sigma_t(\x)}$, $\mu_t(x)$ and $\sigma_t(x)$ are given by Eq.~\ref{eq:gp_mean} \& \ref{eq:gp_var}, and $\phi$ and $\Phi$ are the standard normal PDF and CDF. For actions that collect samples at more then one location, the reward of an action $\tilde{R}(b_t, a)$ is the sum of rewards of the locations sampled by that action.

MVI initially favors collecting observations in areas that have high uncertainty due to sampling maxima from the initial uniform GP belief. As observations are collected and uncertainty diminishes in the GP, the sampled maxima converge to the true maximum and reward concentrates locally at this point, encouraging exploitative behavior. This contrasts with the Upper Confidence Bound (UCB) heuristic, which distributes reward proportional to predictive mean $\mu_t(\mathbf{x})$ and weighted variance $\sigma_t(\mathbf{x})$ of the current GP belief model (Eq. \ref{eq:gp_mean} \& \ref{eq:gp_var}):
$\tilde{R}_{\text{UCB}}(b_t, \mathbf{x}) = \mu_t(\mathbf{x}) + \sqrt{\beta_t}\sigma(\mathbf{x})$. As the robot explores, UCB reward converges to the underlying phenomenon, $f$. The difference in convergence characteristics between MVI and UCB can be observed in Fig.~\ref{fig:figure3_mvi_conv}. 

\section{Experiments and Results}
\label{sec:experiments}
\begin{table*}[tbh!]
  \centering
  \caption{Accumulated True MSS Reward (Eq.~\ref{eq:true_reward}), RMSE, and $\x^*$ Error, Reported as Median (Interquartile Range). \hspace{10cm} Asterisks denote baselines whose difference in performance is statistically significant compared to PLUMES.}
  \begin{tabular}{|l | a|c|c | b | a|c|c | b |}
  \hline
     & \multicolumn{3}{c|}{Convex Simulation Trials} & \multicolumn{1}{b|}{ASV Trial} & \multicolumn{3}{c|}{Non-convex Simulation Trials} & \multicolumn{1}{b|}{Dubins Car Trials} \\
     & \multicolumn{3}{c|}{$\epsilon$ = \SI{1.5}{\meter}, 50 trials} & \multicolumn{1}{b|}{$\epsilon$ = \SI{10}{\meter}, 1 trial} & \multicolumn{3}{c|}{$\epsilon$ = \SI{1.5}{\meter}, 50 trials} & \multicolumn{1}{b|}{$\epsilon$ = \SI{1.5}{\meter}, 5 trials} \\
    \hline
     & \textbf{MSS Reward} & RMSE & $\x^*$ Error & \textbf{MSS Reward} & \textbf{MSS Reward} & RMSE & $\x^*$ Error & \textbf{MSS Reward} \\
    \hline
    PLUMES & \textbf{199} (89) & 3.8 (9.2) & 0.21 (0.23) & \textbf{524} & \textbf{206} (100) & 3.6 (2.1) & 0.25 (0.56) & \textbf{159} (74) \\
    UCB-MCTS & 171 (179)\textbf{*} & 3.7 (9.6) & 0.24 (0.29) & - & 115 (184)\textbf{*} & 3.6 (1.5) & 0.27 (1.18) & 52 (17) \\
    UCB-Myopic & 148 (199)\textbf{*} & 3.6 (9.2) & 0.33 (3.25) & - & 86 (102)\textbf{*} & 3.4 (1.0) & 0.23 (0.34) & 42 (66) \\
    Boustro. & 27 (3)\textbf{*} & 2.7 (10.4) & 0.26 (0.46) & 63 & - & - & -  & - \\
    \hline
  \end{tabular}
  \label{tab:reward_table}
\end{table*}

We analyze the empirical performance of PLUMES in a breadth of MSS scenarios that feature convex and non-convex environments. We compare against three baselines used in environmental surveying: non-adaptive lawnmower-coverage (Boustro., an abbreviation of boustrophedonic \cite{choset1998coverage}), greedy myopic planning with UCB reward (UCB-Myopic) \cite{Sun2017}, and nonmyopic planning with traditional MCTS \cite{browne2012survey} that uses the maximum-likelihood observation and UCB reward (UCB-MCTS) \cite{Marchant2014a}. The performance of UCB planners has been shown to be sensitive with respect to $\beta$ value \cite{Marchant2014a}. In order to avoid subjective tuning, we select a time-varying $\beta_t$ that is known to enable no-regret UCB planning \cite{Srinivas2012, Sun2017}. PLUMES uses continuous-observation MCTS with hyperparameters presented in Auger et al. \cite{auger2013continuous}.

To evaluate the mission performance of all planners, we report accumulated MSS reward (Eq.~\ref{eq:true_reward}), which directly corresponds to the number of scientifically valuable samples collected within an $\epsilon$-ball of the true maximum. This metric is reported for all trial scenarios in Table~\ref{tab:reward_table}. We additionally report several metrics commonly used in IPP to evaluate posterior model quality: overall environmental posterior root mean-squared error (RMSE) and error in posterior prediction of $\x^*$ at the end of a mission ($\x^*$ error). We use a Mann-Whitney U non-parametric significance test \cite{mann1947test} to report statistical significance (p = 0.05 level) in performance between PLUMES and baseline algorithms.

\begin{figure}[th!]
\vspace{10pt}
\centering
\includegraphics[width=0.95\linewidth]{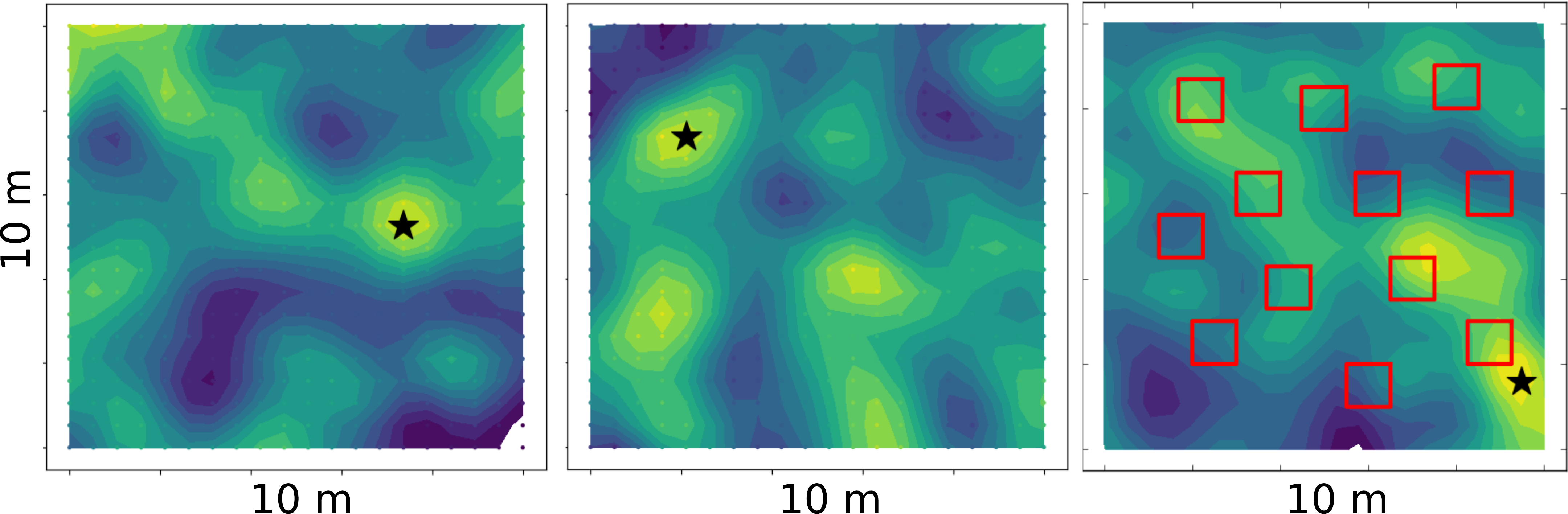}\hfill
\caption{\textbf{Simulation Environments:} The multimodal simulated $10$m $\times$ $10$m environments. Yellow regions are high-valued; blue regions are low-valued. The global maximum is marked with a star. The left and center environments represent convex-worlds (Section \ref{sec:reef}), while the right environment is representative of a non-convex world (Section \ref{sec:car}). \vspace{-0.6cm}}
\label{fig:figure4_freeworld}
\end{figure}

\subsection{Bounded Convex Environments}
\label{sec:reef}

\begin{figure}[b!]
\centering
\vspace{10pt}
\includegraphics[width=0.9\linewidth]{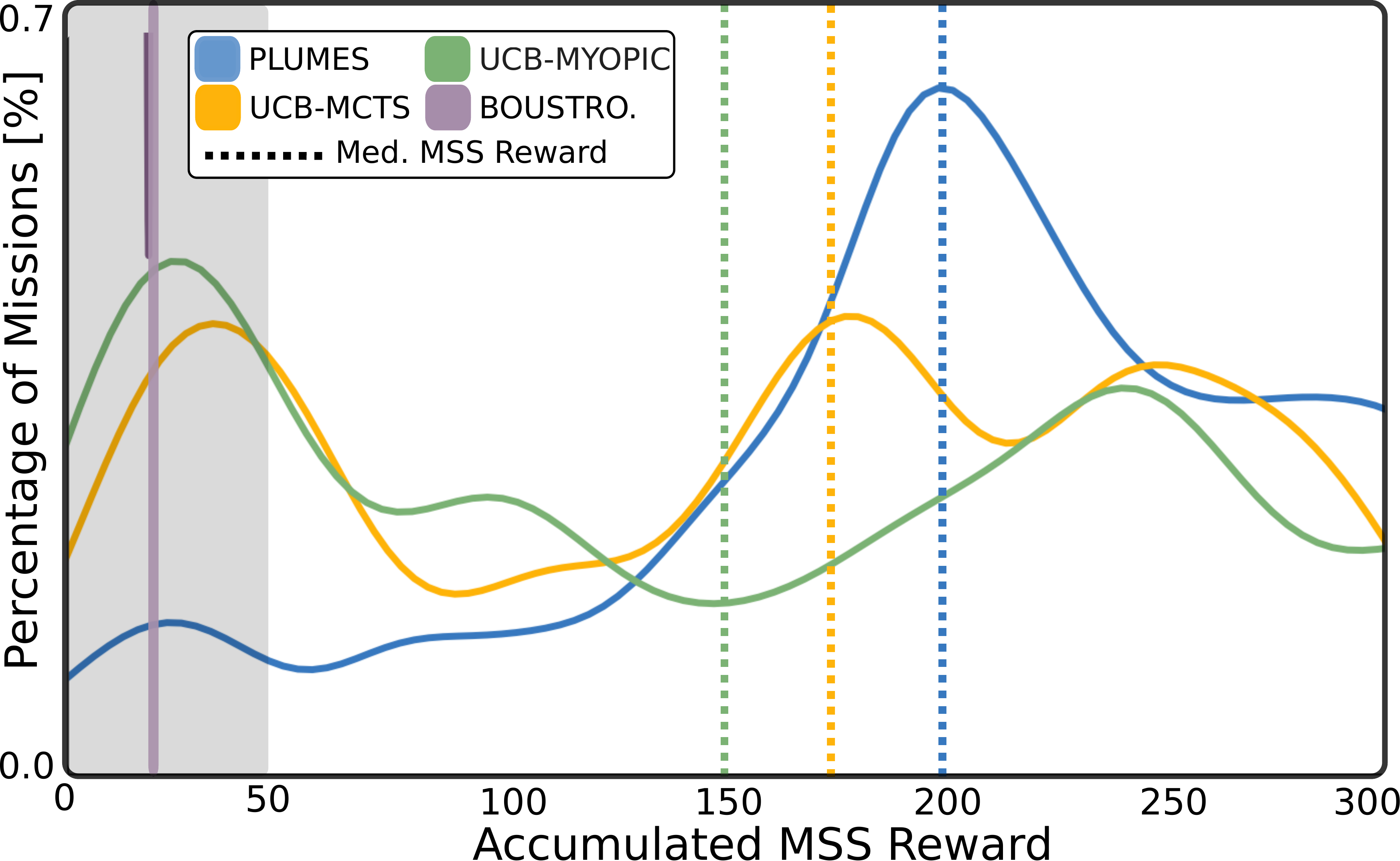}
\caption{\textbf{Distribution of accumulated MSS reward in 50 convex-world simulations:} Accumulated MSS reward is calculated for each trial and the distribution for each planner is plotted as a kernel density estimate (solid line). The dashed lines represent the median accumulated reward for each planner (reported in Table~\ref{tab:reward_table}). The gray area of the plot indicates a low performance region where the planner collected \textless50 samples near the maximum. PLUMES has a single mode near 200, whereas both UCB-based methods are multi-modal, with modes in the low performance region.}
\label{fig:figure5_hist}
\end{figure}

In marine and atmospheric applications, MSS often occurs in a geographically bounded, obstacle-free environment. In 50 simulated trials, we applied PLUMES and our baseline planners to a point robot in a \SI{10}{\meter} $\times$ \SI{10}{\meter} multimodal environment drawn randomly from a GP prior with a squared-exponential covariance function and zero mean ($l = 1.0$, $\sigma^2 = 100.0$, $\sigma_n^2 = 1.0$ [1\%]) (see Fig.\ref{fig:figure4_freeworld}). The action set consisted of ten viable trajectories centered at the robot's pose with path length \SI{1.5}{\meter}, and samples were collected every \SI{0.5}{\meter} of travel. Mission lengths were budgeted to be \SI{200}{\meter}. Nonmyopic planners rolled out to a 5-action horizon and were allowed 250 rollouts per planning iteration. Summary simulation results are presented in Table \ref{tab:reward_table}.

In these trials, PLUMES accumulated significantly (0.05-level) more reward than baselines. The distribution of accumulated reward (Fig.~\ref{fig:figure5_hist}) shows that PLUMES has a single dominating mode near reward 200 and few low-performing missions (reward \textless50). In contrast, both UCB-based methods have distributions which are multimodal, with non-trivial modes in the low-performance region. Boustro. collected consistently few scientifically valuable samples. In addition to collecting many more samples at the maximum, PLUMES achieved statistically indistinguishable levels of posterior RMSE and $\x^*$ error compared to baselines (Table~\ref{tab:reward_table}).

The corresponding field trial for convex-world maximum-search was performed in the Bellairs Fringing Reef, Barbados by a custom-built autonomous surface vehicle (ASV) with the objective of localizing the most exposed coral head. Coral head exposure is used to select vantage points for coral imaging \cite{manjanna2016efficient} and in ultraviolet radiation studies on coral organisms \cite{banaszak2009effects}. Due to time and resource constraints, only one trial of two planners was feasible on the physical reef; we elected to demonstrate PLUMES and Boustro., one of the most canonical surveying strategies in marine sciences.

\begin{figure}[b!]
\centering
\includegraphics[width=\linewidth]{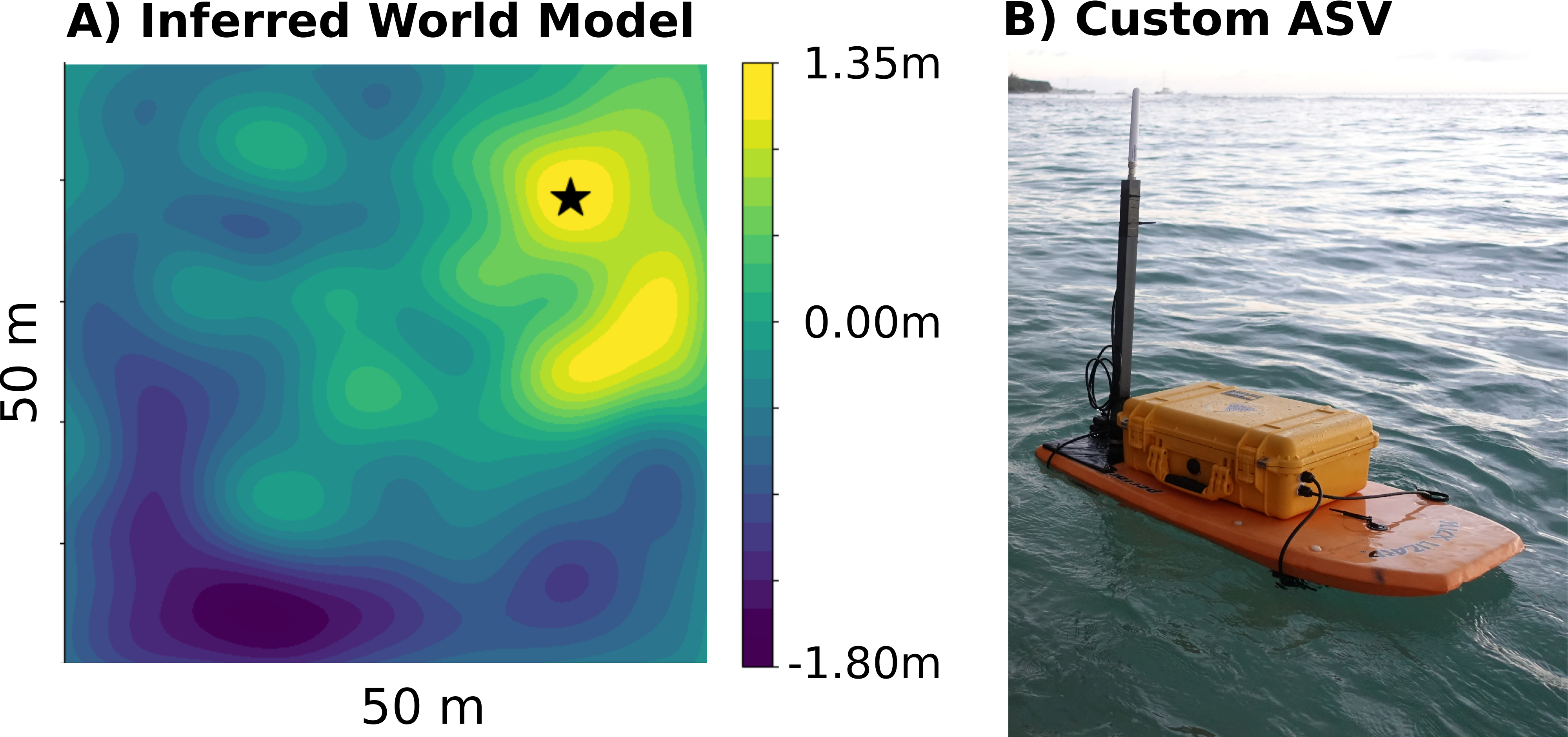}
\caption{\textbf{Coral head map and ASV:} (A) The ground truth bathymetric map inferred from all collected data, mean corrected in depth. Yellow represents shallower depths, and blue is deeper. The global maximum is marked with a black star. (B) The custom ASV used to traverse the \SI{2500}{\meter\squared} region.}
\label{fig:figure6_sl_trials}
\end{figure}

The ASV (\SI{1}{\meter} $\times$ \SI{0.5}{\meter}) had holonomic dynamics and a downward-facing acoustic point altimeter (Tritech Micron Echosounder) with returns at \SI{1}{\hertz}. Ten dynamically-feasible \SI{10}{\meter} straight paths radiating from the location of the ASV were used in the action set. The environment was bounded by a \SI{50}{\meter} by \SI{50}{\meter} geofence. Localization and control was provided by a PixHawk Autopilot with GPS and internal IMU; the fused state estimate was empirically suitable for the desired maximum localization accuracy ($\epsilon$ = \SI{10}{\meter}). The budget for each mission was \SI{1000}{\meter}, which took approx. 45 minutes to travel. The GP kernel was trained on altimeter data from a dense data collection deployment the day before (parameters $l = 2.01$, $\sigma^2 = 0.53$, $\sigma_n^2 = 0.02$ [26\%]). Note the high noise in the inferred GP model, as well as the relatively small length-scale in the \SI{2500}{\meter\squared} field site. The reconstructed bathymetry and vehicle are shown in Fig.~\ref{fig:figure6_sl_trials}.

\begin{figure}[tbh!]
\centering
\includegraphics[width=0.75\linewidth]{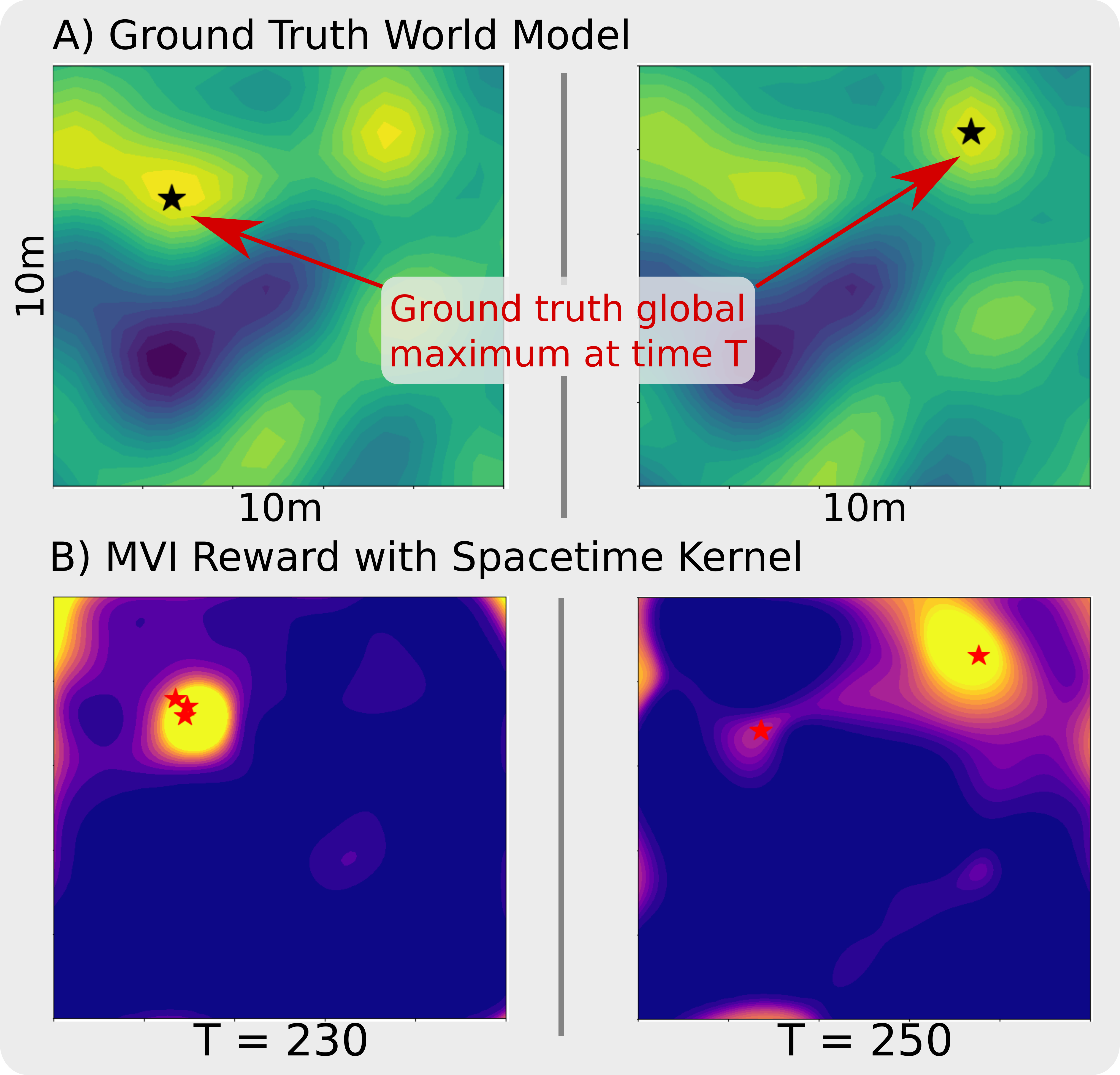}\hfill
\caption{\textbf{Extending PLUMES for Spatiotemporal Monitoring:} (A) The ground truth map at two planning iterations for a dynamic environment. The maximum is marked with a black star, and migrates from the top left to the top right of the world. (B) MVI reward is redistributed by using a spacetime kernel within PLUMES that captures the environment's dynamics.}
\label{fig:figure7_spacetime}
\vspace{-0.5cm}
\end{figure}

PLUMES successfully identified the same coral head to be maximal as that inferred from the GP trained on prior dense data collection, as indicated by accumulated reward in Table~\ref{tab:reward_table}, overcoming the challenges of moving in ocean waves, noisy altimeter measurements, and highly multimodal environment. Additionally, the posterior prediction of $\x^*$ had an error of only \SI{1.78}{\meter} while Boustro. reported \SI{8.75}{\meter} error due to its non-adaptive sampling strategy.

In the Bellairs Fringing Reef trials, the environment was assumed to be static. However, in many marine domains the impact of sediment transport, waves, and tides could physically change the location of a maximum over the course of a mission. PLUMES can be extended to dynamic environments by employing a spatiotemporal kernel in the GP model, which allows for the predictive mean and variance to change temporally \cite{singh2010modeling}. If the dynamics of an environment can be encoded in the kernel function, no other changes to PLUMES are necessary; MVI will be distributed according to the time dynamic. Fig.~\ref{fig:figure7_spacetime} demonstrates the properties of PLUMES with a squared-exponential kernel over space ($l = 1.5$, $\sigma^2 = 100$, $\sigma_n^2 = 0.5$) and time ($l = 100$, $\sigma^2 = 100$, $\sigma_n^2 = 0.5$). In this illustrative scenario, the global maximum moved between planning iteration $T = 230$ and $T = 250$. PLUMES with a spatiotemporal kernel maintained multiple hypotheses about the maximum's location given the random-walk dynamic of the environment, resulting in MVI reward being re-distributed between the two maxima over time.

\subsection{Non-Convex Environments}
\label{sec:car}
We next consider non-convex environments with potentially unknown obstacles, a situation that occurs frequently in practical MSS applications with geographical no-go zones for rover or ASV missions, and in indoor or urban settings. We evaluated PLUMES, UCB-Myopic, and UCB-MCTS planners in 50 simulated trials with the same environments, vehicle, and actions as described in Section \ref{sec:reef}, with the inclusion of 12 block obstacles placed uniformly around the world in known locations (see Fig.\ref{fig:figure4_freeworld}). Boustro. was not used as a baseline because of non-generality of the offline approach to unknown obstacle maps. 

\begin{figure}[b!]
\centering
\vspace{0.15cm}
\includegraphics[width=0.9\linewidth]{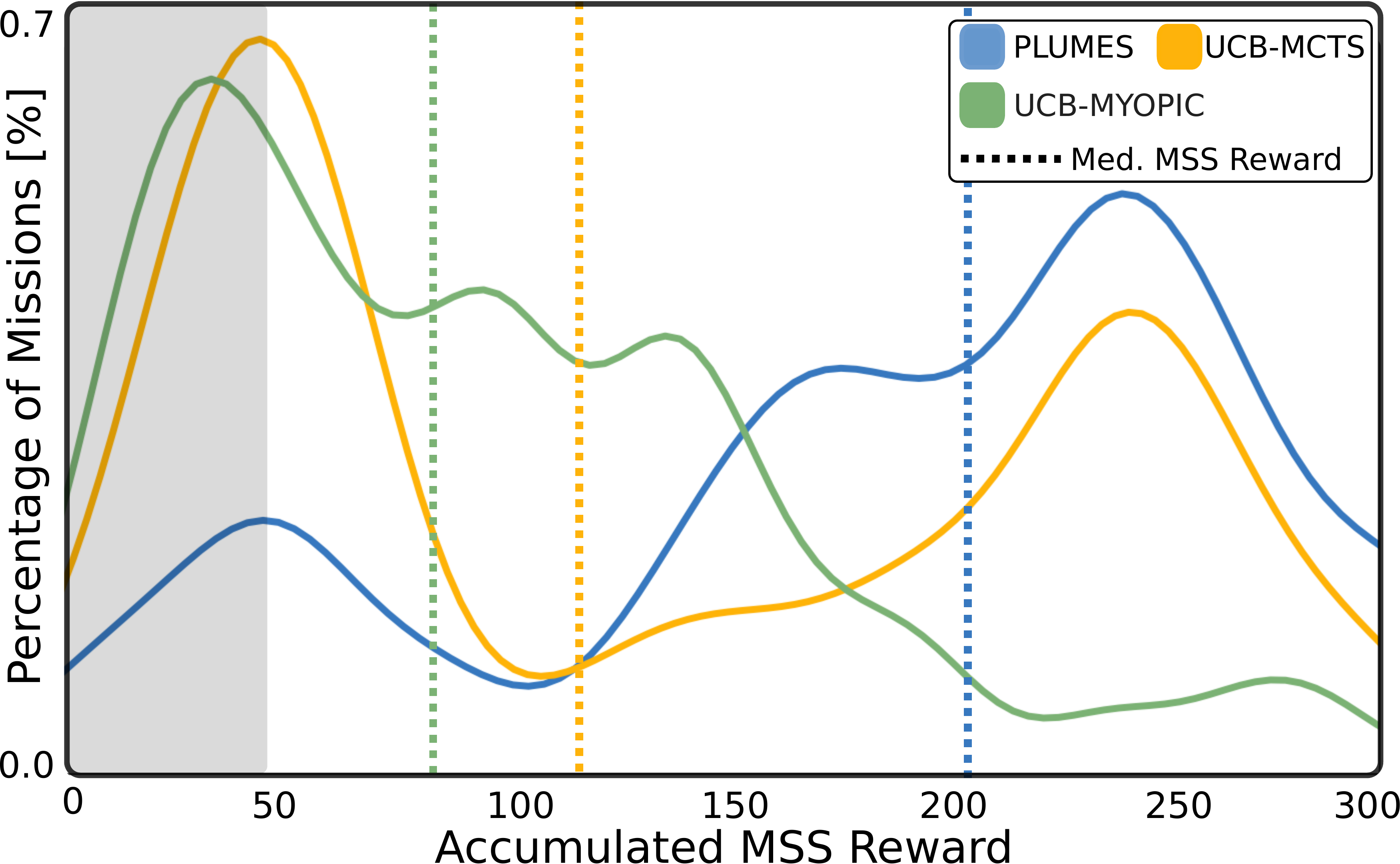}
\caption{\textbf{Distribution of accumulated MSS reward in 50 non-convex mission simulations:} Accumulated MSS reward distribution (solid line) and median (dashed line, reported in Table~\ref{tab:reward_table}) for each planner. The gray area of the plot indicates a low performance region (reward \textless50). PLUMES has few low-performing missions and a primary mode near reward 250. The primary mode of both UCB-based methods is in the low performance region due to convergence to suboptimal local maxima.}
\vspace{-0.25cm}
\label{fig:figure8_hist_clut}
\end{figure}

As indicated in Table~\ref{tab:reward_table}, PLUMES accumulated significantly more MSS reward than UCB-MCTS and UCB-Myopic, at the 0.05-level. The distribution of reward across the trials is visualized in Fig.~\ref{fig:figure8_hist_clut}. Like in the convex-world, the PLUMES has a primary mode between reward 200-250, while the UCB-based planners have a primary mode in the low-performance region (reward \textless50). There was no significant difference between planners with respect to RMSE or $\x^*$ error. 
The fact that PLUMES maximized the true MSS reward while achieving statistically indistinguishable error highlights the difference in exploitation efficiency between PLUMES and UCB-based methods.

\begin{figure}[hbt]
\centering
\vspace{0.15cm}
\includegraphics[width=0.95\linewidth]{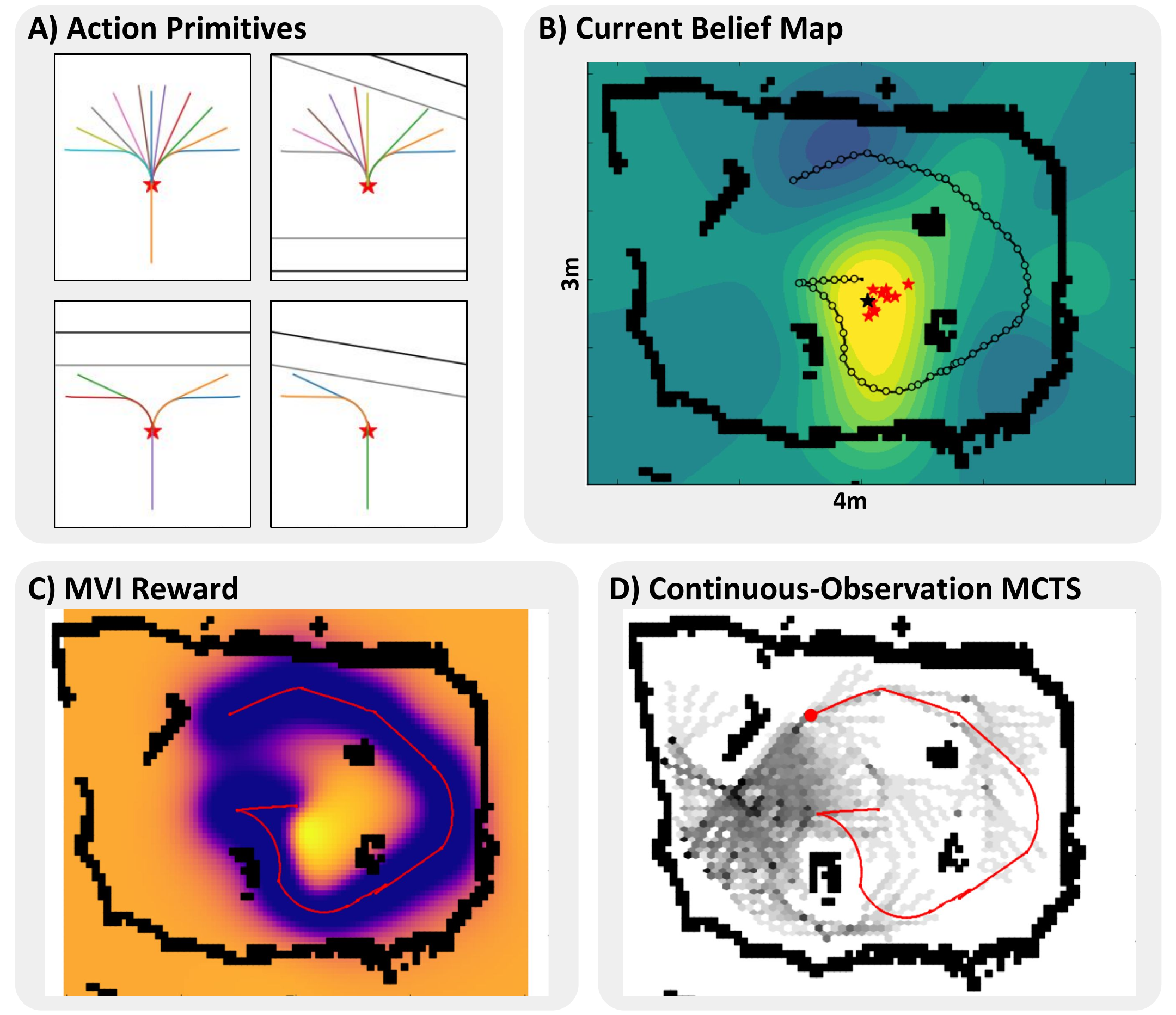}
\caption{\textbf{Snapshot of unknown non-convex map scenario:} (A) shows examples of how the action-primitives change based upon obstacle detection (black lines) and safety padding (grey lines). (B-D) show a planning iteration of PLUMES, starting with the current belief map and obstacle detections (B). The MVI heuristic is illustrated in (C) where lighter regions are higher value. (D) shows the rollout visibility of continuous-observation MCTS where darker regions are visited more often. Areas of high reward are generally visited more often by the search as the tree expands. \vspace{-0.5cm}}
\label{fig:figure9_car}
\end{figure}

The simulation experiments assume that a geometric map is known \textit{a priori}. However in practical applications, like indoor gas leak detection, access to a map may be limited or unavailable. We simulate the scenario in which a nonholonomic car equipped with a laser range-finder must build a map online as it seeks the maximum in a cluttered indoor environment (Fig.~\ref{fig:figure9_car}). We generate a simulated chemical phenomenon from a GP ($l = 0.8$, $\sigma^2 = 100.0$, $\sigma_n^2 = 2.0$ [2\%]), and simulate observations at \SI{1}{\hertz}. The action set for the vehicle consists of eleven \SI{1.5}{\meter} Dubins curves projected in front of the vehicle, one straight path behind the vehicle, and a ``stay in place'' action. Results for five trials are shown in Table \ref{tab:reward_table} and illustrate that PLUMES accumulates more MSS reward than baselines, indicating robust performance.

These simulation and robot trials demonstrate the utility of PLUMES compared to canonical and state-of-the-art baselines in a diverse set of environments with challenging practical conditions. For high-stakes scientific deployments, the consistent convergence and sampling performance of PLUMES is critical and beneficial.

\section{Discussion and Future Work}
\label{sec:discussion}
Online planning methods for robotic maximum seek-and-sample are critical in a variety of contexts, including general environmental monitoring (scientific inquiry, reconnaissance) and disaster response (oil spill, gas leak, radiation). For partially observable environments that can be modelled using a GP, PLUMES is a novel approach for global maximum seek-and-sample that provides several key insights.

This work presents MVI as an empirically suitable alternative to the canonical GP-UCB heuristic in MSS solvers, which is both naturally adaptive and avoids a hand-tuned parameter to balance exploration and exploitation. MVI samples potential global maxima from the robot's full belief state to manage exploration and exploitation. In contrast, heuristic functions like UCB place reward on all high-valued or highly uncertain regions, leading to unnecessary exploration and limiting the time available to exploit knowledge of the true maximum. Ultimately, the MVI heuristic allows PLUMES to collect exploitative samples, while still achieving the same overall level of posterior model accuracy (shown by RMSE) as UCB-based planners. Additionally, continuous-observation MCTS allows PLUMES to search over belief-spaces on continuous functions without discretization or maximum-likelihood assumptions.

One important area of future work for PLUMES is online GP kernel hyperparameter learning \cite{ranganathan2011online}, which is important when only one mission is possible and there is insufficient prior knowledge for hyperparameter selection. Another avenue of future work could be to examine the proprieties of the maxima sampled by MVI, to be used as a heuristic for meta-behavior transitions (e.g., action model switching, dynamic horizon setting) or mission termination. Finally, the performance of PLUMES in non-convex environments is impacted by the chosen discrete action set. Extending PLUMES to continuous actions spaces, in the spirit of, e.g., Morere et al. \cite{morere2018continuous}, would allow increased flexibility in these environments.

\section{Conclusion}
This paper formalizes the maximum-seek-and-sample POMDP and presents PLUMES, an adaptive planning algorithm that employs continuous-observation MCTS and maximum-value information reward to perform efficient maximum-seeking in partially observable, continuous environments. PLUMES outperforms canonical coverage and UCB-based state-of-the-art methods with statistical significance in challenging simulated and real-world conditions (e.g. multiple local maxima, unknown obstacles, sensor noise). Maximum seek-and-sample is a critical task in environmental monitoring for which PLUMES, with theoretical convergence guarantees, strong empirical performance, and robustness under real-world conditions, is well-suited.

\section*{Acknowledgment}
We would like to thank our reviewers for their feedback on this manuscript. Additionally, we thank the SLI group, RRG and WARPlab for their insight and support. This project was supported by an NSF-GRFP award (G.F.), NDSEG Fellowship award (V.P.), and NSF NRI Award 1734400.


\bibliographystyle{IEEEtranBST/IEEEtran}
\bibliography{ipp}

\end{document}